\documentclass[english,12pt]{article}
\usepackage[textwidth=1.75in, textsize=footnotesize]{todonotes}
\presetkeys{todonotes}{fancyline, color=white}{}
\usepackage[
  paperheight=11in,
  paperwidth=8.5in,
  top=0.65in,
  bottom=0.8in,
  right=1.5in,
  left=1.5in
]{geometry}
\usepackage{lmodern}
\usepackage{amssymb,amsmath}

\usepackage{amsthm}
\begingroup
    \makeatletter
    \@for\theoremstyle:=definition,remark,plain\do{%
        \expandafter\g@addto@macro\csname th@\theoremstyle\endcsname{%
            \addtolength\thm@preskip\parskip
            }%
        }
\endgroup

\usepackage{thmtools}

\usepackage{ifxetex,ifluatex}
\ifnum 0\ifxetex 1\fi\ifluatex 1\fi=0 
  \usepackage[T1]{fontenc}
  \usepackage[utf8]{inputenc}
\else 
  \ifxetex
    \usepackage{mathspec}
  \else
    \usepackage{fontspec}
  \fi
  \defaultfontfeatures{Ligatures=TeX,Scale=MatchLowercase}
\fi
\IfFileExists{upquote.sty}{\usepackage{upquote}}{}
\IfFileExists{microtype.sty}{%
\usepackage{microtype}
\UseMicrotypeSet[protrusion]{basicmath} 
}{}
\usepackage{hyperref}
\hypersetup{unicode=true,
            pdftitle={Plausibility and probability in deductive reasoning},
            pdfborder={0 0 0},
            breaklinks=true}
\ifnum 0\ifxetex 1\fi\ifluatex 1\fi=0 
  \usepackage[shorthands=off,main=english]{babel}
\else
  \usepackage{polyglossia}
  \setmainlanguage[]{english}
\fi

\usepackage[parfill]{parskip}
\ifx\paragraph\undefined\else
\let\oldparagraph\paragraph
\renewcommand{\paragraph}[1]{\oldparagraph{#1}\mbox{}}
\fi
\ifx\subparagraph\undefined\else
\let\oldsubparagraph\subparagraph
\renewcommand{\subparagraph}[1]{\oldsubparagraph{#1}\mbox{}}
\fi

\usepackage{csquotes}
\usepackage{epigraph, varwidth}
\usepackage{datetime}
\usepackage[affil-it]{authblk}

\renewcommand{\epigraphsize}{\small}
\renewcommand{\textflush}{flushright}
\renewcommand{\sourceflush}{flushright}
\newcommand{\epitextfont}{\itshape}
\newcommand{\episourcefont}{\scshape}

\makeatletter
\newsavebox{\epi@textbox}
\newsavebox{\epi@sourcebox}
\newlength\epi@finalwidth
\renewcommand{\epigraph}[2]{%
  \vspace{\beforeepigraphskip}
  {\epigraphsize\begin{\epigraphflush}
   \epi@finalwidth=\z@
   \sbox\epi@textbox{%
     \varwidth{\epigraphwidth}
     \begin{\textflush}\epitextfont#1\end{\textflush}
     \endvarwidth
   }%
   \epi@finalwidth=\wd\epi@textbox
   \sbox\epi@sourcebox{%
     \varwidth{\epigraphwidth}
     \begin{\sourceflush}\episourcefont#2\end{\sourceflush}%
     \endvarwidth
   }%
   \ifdim\wd\epi@sourcebox>\epi@finalwidth 
     \epi@finalwidth=\wd\epi@sourcebox
   \fi
   \leavevmode\vbox{
     \hb@xt@\epi@finalwidth{\hfil\box\epi@textbox}
     \vskip1.75ex
     \hrule height \epigraphrule
     \vskip.75ex
     \hb@xt@\epi@finalwidth{\hfil\box\epi@sourcebox}
   }%
   \end{\epigraphflush}
   \vspace{\afterepigraphskip}}}
\makeatother

\newcommand{\LT}[1]{\prec_{\scriptscriptstyle{#1}}}

\newcommand{\LTD}[2]{\prec_{\scriptscriptstyle{(#1, #2)}}}

\DeclareMathOperator*{\argmin}{arg\,min}

\begin{document}

\title{\large Plausibility and probability in deductive reasoning}
\author{\small Andrew MacFie%
\thanks{
  School of Mathematics and Statistics, Carleton University.
  This work was done in part while the author was a visitor at
  the Harvard John A. Paulson School of Engineering and Applied Sciences.
}}
\newdateformat{amdate}{\THEYEAR.\shortmonthname[\THEMONTH].\twodigit{\THEDAY}}
\date{}

\maketitle

\begin{abstract}
\footnotesize
We consider the problem of rational uncertainty about unproven mathematical
statements, remarked on by Gödel and others.
Using Bayesian-inspired arguments we build a normative model of fair bets
under deductive uncertainty which draws from both probability and the theory of
algorithms.
We comment on connections to Zeilberger's notion of ``semi-rigorous proofs'',
particularly that inherent subjectivity would be present.
We also discuss a financial view with models of arbitrage where traders have
limited computational resources.
\end{abstract}

\tableofcontents

\section{A natural problem: Quantified deductive uncertainty}\label{quantified-deductive-uncertainty}

\subsection{The phenomenon}\label{sec:phenom}

Epistemic uncertainty is usually defined as uncertainty among physical
states due to lack of data or information. By information we mean facts
which we observe from the external world. For example, whether it rains
tomorrow is a piece of information we have not observed, so we are
uncertain about its truth value. However, we also may have uncertainty
about purely deductive statements, which are completely determined by
the information we have, due to limited reasoning ability. That is,
before we have proved or refuted a mathematical statement, we have some
\emph{deductive uncertainty} about whether there is a proof or
refutation.

Under deductive uncertainty, there is a familiar process of appraising a
degree of belief one way or the other, saying a statement has high or
low plausibility. We may express rough confidence levels in notable open
conjectures such as \(\mathbf{P} \neq \mathbf{NP}\) \cite{sciencase}
or the Goldbach conjecture, and we also deal with plausibility in
everyday mathematical reasoning. Sometimes general patterns show up
across problems and we extrapolate them to new ones. If we have an
algorithm and are told it runs in time \(O(n \lg n)\), we usually assume
that this implies good practical performance because this is a commonly
observed co-occurrence. So the plausibility of the running time being
\(10^{10!} n \lceil \lg n \rceil\) is considered particularly low. Any
mathematical result seen as ``surprising'' must have been a priori
implausible. Et cetera. Many more examples of plausibility in
mathematics may be found in \cite{polya,mazur2014plausible}.

In some instances it may be natural to
quantify deductive uncertainty, and perhaps speak of ``probabilities''.
For example, let \(d\) be the \(10^{100}\)th decimal digit of \(\pi\).
If we have not computed \(d\) and all we know is that, say, \(d\) is odd, it
feels like \(d\) has a uniform probability distribution over
\(\{1,3,5,7,9\}\). Mazur \cite[Sec.~2]{mazur2014plausible} would
describe this as an application of the principle of insufficient reason.
We use the same ``symmetry'' argument to state the probability that a
given number \(n\) is prime via the prime number theorem or Fermat
primality test. Probabilities also show up in enumerative induction,
where confidence in a universal quantification increases as individual
instances are verified. The four color theorem is one of many cases
where only positive instances could be found and eventually a proof was
given. Furthermore, to this theorem and other similar claims there are
relevant 0-1 laws \cite{gao01} which state that the ``conditional
probability'' of a uniform random instance being a counterexample, given
that counterexamples exist, goes to \(1\) asymptotically. With this fact
one can use individual instances to ``update'' a Bayesian probability on
the universal statement. Bayesianism in practical mathematics has been
discussed previously in \cite{corfield2003towards}.

Notwithstanding the examples above, mathematicians generally leave their
uncertainty unquantified. This may be due to haziness about, for
example, what a ``60\% chance'' means, and how probability should be
used, in the context of deductive reasoning. One point to emphasize is
that we are referring to subjective uncertainty rather than any due to
inherent ``randomness'' of mathematics. Of course there is nothing
random about whether a proof exists of a given mathematical statement.
Frege, speaking on mathematical reasoning, appears to note this lack of
randomness as a problem: ``the ground {[}is{]} unfavorable for
induction; for here there is none of that uniformity which in other
fields can give the method a high degree of reliability''
\cite{frege1980foundations}.
I.e.\ a set of propositions may be homogeneous in some ways but their truth
values cannot be seen as merely IID samples because the propositions are
actually distinct and distinguishable.
However, if we model bounded reasoning we may indeed have an analogy to other
forms of uncertainty.

\subsection{The (modeling) problem}\label{the-modeling-problem}

Gödel mentions deductive probabilities in a discussion of empirical
methods in mathematics \cite{gocol}:

\begin{quote}
It is easy to give examples of general propositions about integers where
the probability can be estimated even now. For example, the probability
of the proposition which states that for each \(n\) there is at least
one digit \(\neq 0\) between the \(n\)-th and \(n^2\)-th digits of the
decimal expansion of \(\pi\) converges toward \(1\) as one goes on
verifying it for greater and greater \(n\).
\end{quote}

In commentary, Boolos naturally asks how such probabilities would be
computed \cite{gocolboolos}:

\begin{quote}
One may, however, be uncertain whether it makes sense to ask what the
probability is of that general statement, given that it has not been
falsified below \(n = 1000000\), or to ask for which \(n\) the
probability would exceed \(.999\).
\end{quote}

With Boolos, we want to know, how would subjective deductive
probabilities work in general? Are there right and wrong ways to assign
these probabilities? Do they even make sense? These questions have both
positive and normative versions; we focus on the normative.

Bayesianism is the theory of probabilities for physical uncertainty
\cite{weisberg2011varieties}. It gives an interpretation of
probability, where we take a probability space, and interpret the
probability measure as assigning subjective degrees of belief to events
which represent expressible physical states. Looking from the reverse
direction, Bayesianism argues from the nature of physical uncertainty to
reach a conclusion that degrees of belief should form a probability
space.
In this second sense we can think of Bayesianism as a proof, where the
premise is physical uncertainty, the inferences are rationality arguments,
and the conclusion is probability theory.
There are two ways to make use of a proof to learn something new.
First, we can apply the theorem if we are able to satisfy the premise.
Here this would mean reducing deductive uncertainty to physical
uncertainty by defining virtual information states.
This general problem of deductive probabilities has received some attention in
the literature (the sample space is taken to consist of complete formal
theories) as we mention in later sections.
But what if there is no valid way to define virtual information for deductive
uncertainty, i.e.\ what if probability theory is not appropriate for
deductive uncertainty in this manner?
What if something else is?
The second way to learn from a proof is to imitate the proof technique.
Here we would start with the premise of deductive uncertainty, proceed using
analogous but adapted rationality arguments, and
reach a conclusion which is a set of mathematical rules possibly different
from probability theory.
We focus on this approach.

The first step is to fix a concrete and unambiguous way to quantify
uncertainty. If we assign a number to our belief in a statement, what
does that mean? And is it always possible for us to do so? In
Bayesianism, uncertainty is quantified by a single real number from
\([0,1]\) and a prominent operational definition of this quantification
is based on fair bets \cite[Ch. 13]{Russell:2009:AIM:1671238},
\cite[Sec. 2.2.1]{weisberg2011varieties}. This operationalization appears to work
equally well for deductive uncertainty. That is, anyone can express
their uncertainty about a mathematical statement \(\phi\) using a number
in \([0,1]\) which encodes the betting payoffs they consider acceptable
if they were to bet on \(\phi\). We call these values
\emph{plausibilities}. (This is not to be confused with other usages
such as ``plausibility measures'' \cite[Sec.~2.8]{halpern2005reasoning}).
We assume this operationalization is meaningful and understood.


In the context of physical uncertainty, Bayesianism adds constraints on
what plausibilities/probabilities should be for a rational agent, namely
coherence (so plausibilities are probabilities),
conditionalization, regularity and other guidance for selecting priors
\cite{weisberg2011varieties, kc}.
Also, probabilistic forecasts may be rated on accuracy using loss
functions \cite{lai2011evaluating}.

However, the assumptions of Bayesianism on which these constraints are
based do not necessarily still apply to deductive plausibilities and
indeed we may have additional or different requirements. Thus the
precise question to answer is, what constraints should be put on
deductive plausibilities and what mathematical structure results?

\section{Formal representation of
plausibilities}\label{formal-representation-of-plausibilities}

\subsection{Plausibility functions}\label{plausibility-functions}

Fix an encoding of deductive statements into finite strings so that
there is a decision problem \(\Pi \subseteq \{0,1\}^*\) corresponding to
the true statements. We take an association of plausibilities to encoded
statements as a function \(p: \{0,1\}^* \to [0,1]\). We call \(p\) a
\emph{plausibility function}. A plausibility function represents an
agent's uncertainty about \(\Pi\). One could also have defined a
plausibility function to take a finite sequence of input strings and
return a finite sequence of plausibilities, that is, working at the
level of bet systems instead of bets.

\subsection{Languages}\label{languages}

Finding proofs is a matter of computation, so our reasoning abilities
are equivalent to our computational resources; and generally we will
experience deductive uncertainty when faced with any intractable
computational problem.
Importantly, we cannot meaningfully talk about problems with only one input,
since obviously the best output is the actual truth value.
So we must talk of uncertainty about an entire set of inputs simultaneously.

Probability spaces consist of a measurable space and a probability
measure. In Bayesianism, the measurable space may be substituted by a
``sentence space'' which is closed under logical operations. In the
deductive case, any nontrivial problem \(\Pi\) has an input set that is
trivially closed under logical operations, since any input is logically
equivalent to ``true'' or ``false''.
We conclude that the problem \(\Pi\) need not a priori have any particular
syntactic structure and we may consider standard problems from theoretical
computer science.

There is a line of research on deductive uncertainty where the inputs
come from a first-order language. Typically this work aims at finding
composites of logic and probability theory, and there is less focus on
practicality.
The most recent work is by Garrabrant et al.\
\cite{garrabrant} and \cite[Sec.~1.2]{garrabrant} reviews previous
literature.
In the
present work we instead restrict our attention to decidable problems. We
do this because inconsistent logics do not make sense in terms of
betting. So first-order logics are problematic due to Gödel's first and
second incompleteness theorems.

\subsection{Epistemic quality vs.~computation
costs}\label{epistemic-quality-vs.computation-costs}

For a given problem \(\Pi \subseteq \{0,1\}^*\) we seek an
\emph{epistemic improvement relation} \(\LT{\Pi}\) on plausibility
functions, where \(q \LT{\Pi} p\) iff \(p\) is a strictly better
uncertainty assignment for \(\Pi\) than \(q\), ignoring any
computational costs of the functions.
For example, if we decide to require
regularity, we would say that for all functions \(p\) and \(q\), if
\(p\) is regular and \(q\) is not then \(q \LT{\Pi} p\). Guidance for
selecting plausibility functions is then based on weighing \(\LT{\Pi}\)
against computational costs.
If we are given a probability distribution
on inputs, we take the distributional decision problem
\((\Pi, \mathcal{D})\) and consider a distribution-specific relation
\(\LTD{\Pi}{\mathcal{D}}\). We refer to the relation as an order but it
is not necessarily total.

Improvement relations can be found in other modeling contexts. For
algorithm running time sequences we use asymptotic relations such as big
O or polynomial growth vs.~non-polynomial growth. Fine-grained relations
may be used if the model of computation is fixed. The Nash equilibrium
represents a kind of ordering which expresses that a strategy profile
can be improved from one player's perspective. Pareto optimality is
similar but for group rationality. Another example is the
Marshall/Kaldor-Hicks social welfare improvement relation in economics
\cite{feldman1998kaldor,friedman1986price}. This last relation is
useful even though it cannot be defined to be both transitive and
antisymmetric.

In general we have a tradeoff between epistemic quality (whatever we
determine that to be) and computational complexity. A theory of
deductive uncertainty must not only define gradients of epistemic
quality but dictate how to make this tradeoff. If we allow arbitrary
computations, the problem immediately disappears. E.g.\ there is a
temptation to look into Solomonoff induction \cite{kc} as a model of
inductive reasoning applied to mathematical knowledge. This would be an
attempt to formalize, e.g.~Pólya's patterns of plausible reasoning
\cite{polya,mazur2014plausible}, such as ``\(A\) analogous to
\(B\), \(B\) more credible \(\implies A\) somewhat more credible''.
However we must be cautious, because an incomputable prior cannot be the
correct tradeoff between quality and efficiency.

Computation costs may or may not be measured asymptotically. Asymptotic means
no finite set of inputs can make a difference. If we use asymptotic
complexity this forces us to define \(\LT{\Pi}\) so that it is
compatible, i.e.~also asymptotic. As an example, utility in game theory
is generally not compatible with asymptotic computation costs. There
are, however, game models which trade off running time and other
considerations in a non-trivial way, for example using discounted
utility. In economics and game theory, the concept of ``bounded
rationality'' refers to decision making with limitations on
reasoning/optimization power, such as imperfect recall, time
constraints, etc. \cite{rubinstein1998modeling}. We note some
economic models which incorporate bounded computation: game-playing
Turing machines with bounded state set \cite{megiddo1986play},
automata models \cite{papadimitriou}, machines as strategies and
utility affected by computation cost
\cite{halpern2015algorithmic,fortnow2009bounding}, information
assymetry in finance \cite{arora2010computational}.
If a game model uses a practically
awkward criterion for algorithm performance, simple models of
computation may be used, or equilibria may be reasoned about without
analyzing a particular algorithm.

A simple approach to the tradeoff is to fix a resource bound and
consider as ``feasible'' only functions that can be computed within the
bound. Then, given \(\LT{\Pi}\), we optimize over the subset of feasible
plausibility functions. This is the method we focus on in the remainder.
E.g.\ we may assume the Cobham-Edmonds thesis and consider \(\LT{\Pi}\)
restricted to polynomial-time-computable functions.

We make the assumption that we, as modelers, are always capable of
analyzing given plausibility functions however is necessary to evaluate
\(\LT{\Pi}\) and analyze computational complexity. This is of course not
true, as discussed in
\cite{aumann2005musings,modarres2002methodological} which consider
bounded rationality in economic systems. However this is a worthy
assumption since building meta-uncertainty into the model creates a
regress which would add significant complexity. Thus we can say that
optimizing \(\LT{\Pi}\) is the rational way to select a plausibility
function even if we are not currently able to do so constructively.
Particularly, when we analyze functions according to an input distribution,
the business of defining the distribution is that of the unbounded analyst.
In practice, e.g.~approximation algorithms are analyzed, even if the problem
they attempt to approximate is intractable.

\subsection{Conditional plausibility}\label{conditional-plausibility}

If we select plausibility functions by optimizing \(\LT{\Pi}\) over
feasible functions, the definition of feasibility could change over time
or simply vary across contexts, so in general we speak of
\emph{conditional plausibility} functions \(p(\cdot | S)\), where \(S\)
is an oracle or complexity class or other representation of the resources
available to compute the function. Another interpretation is that, over time,
computation costs may effectively come down, enlarging the budget set of
feasible functions. This notation and terminology indicates an analogy
where knowledge, in the computational sense (roughly that of
\cite[Sec.~9.2.3]{goldreich_book}, \cite[Sec.~7.2]{goldwasser1989knowledge}),
takes the place of information.


In Bayesianism, conditionalization is the process by which updates on
new information must occur. I.e.\ after observing \(A\), our new
probability of \(B\) is \(P(B|A)=P(A \cap B)/P(A)\).
We note that
conditionalization is a uniform process in that there is a finite rule
that performs updates for all events \(A\) at all times. If there is an
infinite set of possible \(S\), we could restrict to uniform-updating
plausibility functions, i.e.~those which take a representation of \(S\)
as a parameter. In, for example, Garrabrant's model \cite{garrabrant},
the plausibility function takes an additional parameter \(n\), which is
the number of stages to run. However this level of analysis is beyond
our scope.

\section{Rational plausibilities in mathematics}\label{rational-plausibilities}

\subsection{Scoring rules}\label{epistemic-quality}


As a method of eliciting quantified uncertainty, fair bet odds are equivalent
to asking ``what is an objective chance \(\bar{p}\) such that your uncertainty
of this event is equivalent to that of an event with objective chance
\(\bar{p}\)?''.
These are also equivalent to strictly proper scoring rules,
which go back to de Finetti and beyond \cite{de1981role}.
This refers to a situation where the agent is asked for a number $x \in [0,1]$
representing uncertainty about a proposition $\phi$, and then the agent
receives a score of $B([\phi], x)$.
Here we use Iverson brackets where $[\phi]$ is $1$ if $\phi$ is true, and
$0$ otherwise.
(The word ``score'' is a bit of a misnomer because lower scores are better.)
The scoring function $B$ is strictly proper iff
\[ \argmin_x y B(1, x) + (1-y)B(0, x) = y, \]
i.e.\ if we assumed $[\phi]$ is a $\textsc{Bernoulli}(y)$-distributed random
variable, then we obtain the optimal expected score by choosing $x = E[\phi] =
y$.
We note that with both betting and scoring, the agent is
presented with a simple decision where the only computational difficulty
comes from the one instance of some decision problem.
This allows us to conclude that their equivalence holds in the computational
setting as well.

Since plausibilities encode how we make simple decisions, more desirable
plausibility values are those that lead to better outcomes from
decisions.
Dutch book arguments justifying Bayesianism are essentially saying if you can
avoid losing money in a bet, you should.
On the other hand, scoring rules are generally considered to index
epistemic quality.
In fact, the concepts of betting and scoring rules are essentially the same
for classical uncertainty.
It is
shown in \cite{predd2009probabilistic} that Dutch books exist iff an
agent's forecasts are strictly dominated, where domination means there is
another plausibility assignment that obtains a better score in all outcomes,
according to a continuous proper scoring rule.
We take this scoring rule characterization of Bayesianism (which led to
probability theory in that case) and apply it to deductive plausibilities
via analysis of plausibility functions.

Proper scoring rules conveniently associate a real number to each of our
plausibilities.
The Brier score
has some desirable properties \cite{selten1998axiomatic}.
The logarithmic score also has desirable properties and is closely related
to information-theoretic entropy.
Given any proper scoring rule, one can always construct a decision
environment such that performance in the environment is precisely performance
according to the scoring rule.
However, scoring rules are equivalent in the context of conditional expectation
\cite{banerjee2005optimality}
and by analogy we may expect rational plausibilities to approximately share
this property and others \cite{cond, resnick}.

Worst-case scoring of plausibility functions leads to trivialities since
\(p\equiv 1/2\) is optimal unless the problem can be exactly solved.
An alternative in some cases would be to consider inputs of length $\leq
n$ rather than $n$, but we focus on the standard practice of considering inputs
of the same length.
In the context of professional mathematics, we may take an average-case
approach with a fixed input distribution.
This is a non-adversarial model where the distribution reflects inputs that
come up in practice and the agent is betting \enquote{against nature}.

If inputs are distributed according to an ensemble \(\mathcal{D}\), we may say
that \(q \LTD{\Pi}{\mathcal{D}} p\) if the expected score of \(p\) is less
than that of \(q\).
The comparison may be asymptotic.
This is the model used in \cite{vadim} which develops some relevant
mathematical theory.
More general background includes average-case complexity theory
\cite{williamssurvey, bogdanov}
and probabilistic numerics
\cite{ritter2007average, cockayne2017bayesian}.
(A very similar view is considering \(p\) as an unnormalized distribution on
strings of length \(n\), and finding the statistical distance between the
normalized distribution and the normalized distribution corresponding to
\(1_\Pi\).
There we would require at least that \(E(p)=E(1_\Pi)\).)

\subsection{\texorpdfstring{Foundations of ``semi-rigorous
proofs''}{Foundations of semi-rigorous proofs}}\label{foundations-of-semi-rigorous-proofs}

Zeilberger presented the concept of a ``semi-rigorous proof''
and predicted that it might become acceptable by the mathematical community:
\enquote{I can envision an abstract of a paper, c.\ 2100, that reads:
\enquote{We show, in a certain precise sense, that the Goldbach conjecture is
true with probability larger than $0.99999$}} \cite{semirigorous}.
In order for such a result to be useful, perhaps there must be cases where
plausibilities are objective.

According to $\LTD{\Pi}{\mathcal{D}}$, are plausibilities objective or
subjective?
Should we expect people to agree on plausibilities?
There are various sources of subjectivity: how to embed individual questions in
problems, first-order logic issues, and problems $\Pi$ where
\(\LTD{\Pi}{\mathcal{D}}\) has no unique optimum.
For example, take Gödel's $\pi$ problem from Sec.~\ref{the-modeling-problem}:
Let $\phi$ represent the sentence, for each \(n\) there is at least
one digit \(\neq 0\) between the \(n\)-th and \(n^2\)-th digits of the
decimal expansion of \(\pi\).
To condition on observed digits of $\pi$, we can allow access to an oracle
that checks $\phi$ for large ranges of the digits of $\pi$.
First, there may not be a proof or refutation of $\phi$, in which case betting
outcomes are undefined.
Second, if $\phi$ is decideable, its truth value can always be hard coded in
$p$ even if we embed $\phi$ in a class of problem instances which is
standard and universal.
Analyzing general methods of computing plausibilities, rather than individual
values, does makes sense since people typically use heuristics in a consistent
manner across problems.

In traditional Bayesianism there is a seemingly ineradicable source of
subjectivity from the choice of prefix Turing machine used to define
Solomonoff’s prior.
Any one string can be assigned (almost) an arbitrary probability.
Perhaps we are left with an analogous but different kind of
subjectivity for mathematical plausibilities.

We stated at the end of Sec.~\ref{sec:phenom} that mathematicians may be
uncomfortable with putting too much focus on plausibilities.
Gödel says,
``I admit that every mathematician has an inborn
abhorrence to giving more than heuristic significance to such inductive
arguments'' \cite{gocol}.
Also, Corfield notes, ``Pólya himself had the intuition
that two mathematicians with apparently similar expertise in a field
might have different degrees of belief in the truth of a result and
treat evidence for that result differently''
\cite{corfield2003towards}.
However, the physics community has had notable success using non-rigorous
mathematical methods.

One practical issue with publishing probabilities for mathematical
problems is error amplification.
If we take the conjunction of two ``independent'' uncertain statements we end
up with uncertainty greater than that of either of the original statements,
which means confidence erodes over time.
In mathematics this is undesirable since we are used to taking arbitrarily long
sequences of conjunctions and implications with no loss of validity.

More on probabilistic proofs is found in \cite{transferability}.
The potential for models of uncertainty in mathematics to explain the use
of large computations to increase confidence in conjectures is noted
in \cite{corfield2003towards}.

\section{Arbitrage pricing in markets with computational constraints}

Here we use the language of finance; see the paper \cite{nau2001finetti} for
some discussion on the equivalence of Dutch books and arbitrage.
Suppose instead of a binary function $1_\Pi$ we estimate a continuous-valued
function $F$.
Consider a $2$-period market where at time $t=0$ the seller(s) offer
a price $f(x)$ for each asset $x \in \{0,1\}^*$.
Also at time $t=0$, a buyer buys a quantity $g(x)$ of the asset $x$ at the
prices given by $f$, where $g(x) <0$ indicates short selling.
At time $t=1$, each asset $x$ has terminal payoff determined by $F(x)$.
Ignoring time discounting, the buyer's gain from $x$ is
\begin{equation*} \label{eq:gain}
F(x)g(x) - f(x)g(x).
\end{equation*}

The function $F$ may be of two possible kinds.
If the payoffs are deterministic, then computation of $F$ is presumably
non-trivial but ultimately must be performed.
On the other hand, the payoffs represented by $F$ may be expected values of
some random variables, in which case there is no need for $F$ to be
computationally tractable, but sampling from the relevant probability
distributions should be.
Roughly speaking, these situations we describe could be found where information
is fixed throughout the market history $t=0,1$ and $F$ is a reference model
used to price the assets $x$, and where $F$ has nontrivial computational
complexity.
In the more general and likely more realistic case where information is not
fixed, there would be a tradeoff between waiting for more information and
performing lengthier computations which is not included in this model.

Computational aspects of economic price models such as Arrow-Debreu
equilibria \cite{takayama1985mathematical} and combinatorial auctions
\cite{cramton2007overview} have received interest in the literature
\cite{deng2008computation,hirsch1987exponential,axtell2005complexity}.
Pricing is often based on probabilistic models.
Computational difficulty leads to the use of Monte Carlo methods e.g.\ for
numerical solutions to stochastic differential equations, Bayesian posteriors,
and other simulations
\cite{sauer2012numerical,gelman1995bayesian,ryza2017advanced}.
Probabilistic inference in Bayesian networks is known to be computationally
hard in general \cite{dagum1993approximating}.
The hardness of pricing arbitrary exotic derivatives is explored in
\cite{braverman2014computational}.

What does arbitrage mean in this context?
Suppose $C^g$ is a class of functions $g$.
Then we would say the prices $f$ admit arbitrage from $C^g$ based on
the buyer's gain
\[b_n(g) = \sum_{x \in \{0,1\}^n} F(x)g(x) - f(x)g(x)\]
for different $g \in C^g$.
In addition to computational constraints, we may restrict $C^g$ to polynomial
growth and polynomial-sized support to reflect practical trading limits.
We could also allow the arbitrageur to randomly generate an asset $x$,
subject to computational constraints, so $b_n$ would be an expected value.
The classical definition of arbitrage \cite{delbaen2006mathematics} suggests
requiring there exist $g \in C^g$ such that
$b_n(g) \geq 0$ for all $n$ and $b_n(g) > 0$ infinitely often.
We say a sequence $a_n$ is negligible iff $a_n = n^{-o(1)}$.
A relaxation of strict arbitrage would require having $b_n(g)$ positive and
non-negligible infinitely often and not having $-b_n(g)$ positive and
non-negligible infinitely often.

In current complexity theory, it is not clear whether this form of arbitrage
can or cannot be avoided.
The paper \cite{gutfreund2007if} shows that for the SAT problem, if $f$ is
computable in polynomial time and $C^g$ is the set of functions computed in
polynomial time, then we can always get $b_n > K$ infinitely often for some
fixed $K>0$ unless \(\mathbf{P} = \mathbf{NP}\).


\section{Conclusion}

Even within traditional Bayesianism, the assumption of computational
unboundedness can be undesirable; this is known as the problem of logical
omniscience
\cite{sep-logic-epistemic,sep-epistemology-bayesian,halpern2011dealing}.
Some work has been done on formal models for logical non-omniscience,
including a resource-bounded AIXI \cite{hutterbook} and resource-bounded
Solomonoff prior \cite{kc}, although these are somewhat ad-hoc.
As mentioned in the previous section, in general an agent must make decisions
based on the available information and the computational costs of reasoning.
A powerful model is given in \cite{orseau2012space}, although
this model may be hard to integrate with computational complexity theory.
Ideally a general decision theory
would build on probability and computational complexity in a way
that allows us to exploit our mathematical understanding of those fields.
Still, even if we have a definition of the optimal decision rule, there
may be research remaining in algorithm design and analysis to actually
construct it.
And that research in turn may utilize a Bayesian perspective as in the
pseudo-distributions of the sum-of-squares algorithm \cite{bayes2,
barak2016nearly}.

\vspace{2\baselineskip}
\textbf{Acknowledgements.}
The author thanks
Boaz Barak for discussions, and also
Zhicheng Gao,
Vadim Kosoy,
and Corey Yanofsky.

\addcontentsline{toc}{section}{References}
\bibliography{divingboard1}
\bibliographystyle{plain}

\end{document}